\documentclass[conference]{IEEEtran}
\IEEEoverridecommandlockouts

\usepackage{cite}
\usepackage{amsmath,amssymb,amsfonts}
\usepackage{algorithmic}
\usepackage{graphicx}
\usepackage{textcomp}
\usepackage{bbding}
\usepackage{multirow}
\usepackage{xcolor}
\def\BibTeX{{\rm B\kern-.05em{\sc i\kern-.025em b}\kern-.08em
    T\kern-.1667em\lower.7ex\hbox{E}\kern-.125emX}}
\begin{document}

\title{Image Quality Assessment With Compressed Sampling}

\author{\IEEEauthorblockN{Ronghua Liao, Chen Hui, Lang Yuan, Haiqi Zhu, Feng Jiang}
\IEEEauthorblockA{
\textit{Harbin Institute of Technology, }Harbin, China}}

% \author{\IEEEauthorblockN{1\textsuperscript{st} Ronghua Liao}
% \IEEEauthorblockA{\textit{Harbin Institute of Technology}\\
% China \\
% email address or ORCID}
% \and
% \IEEEauthorblockN{2\textsuperscript{nd} Chen Hui}
% \IEEEauthorblockA{\textit{Harbin Institute of Technology}\\
% China \\
% email address or ORCID}
% \and
% \IEEEauthorblockN{3\textsuperscript{nd} Lang Yuan}
% \IEEEauthorblockA{\textit{Harbin Institute of Technology}\\
% China \\
% email address or ORCID}
% \and
% \IEEEauthorblockN{4\textsuperscript{nd} Feng Jiang}
% \IEEEauthorblockA{\textit{Harbin Institute of Technology}\\
% China \\
% email address or ORCID}
% }

\maketitle

\begin{abstract}
No-Reference Image Quality Assessment (NR-IQA) aims at estimating image quality in accordance with subjective human perception. However, most methods focus on exploring increasingly complex networks to improve the final performance, accompanied by limitations on input images. Especially when applied to high-resolution (HR) images, these methods offen have to adjust the size of original image to meet model input. To further alleviate the aforementioned issue, we propose two networks for NR-IQA with Compressive Sampling (dubbed CL-IQA and CS-IQA). They consist of four components: (1) The Compressed Sampling Module (CSM) to sample the image (2) The Adaptive Embedding Module (AEM). The measurements are embedded by AEM to extract high-level features. (3) The Vision Transformer and Scale Swin TranBlocksformer Moudle (SSTM) to extract deep features. (4) The Dual Branch (DB) to get final quality score. Experiments show that our proposed methods outperform other methods on various datasets with less data usage.
\end{abstract}


\section{Introduction}
While utilizing images from the internet, we should be mindful of their potential transformation journey. These images might have been captured by a camera, compressed for sharing online and disseminated widely before finally reaching us. These low-quality images can affect the our visual feelings and even cause deadly problems, especially in autonomous driving. For the reason, it is crucial to predict the perceptual image quality in our daily lives.

Objective Image Quality Assessment (IQA) is an approach that uses computational models to ascertain the perceived quality of an image from a human perspective. The objective quality metrics can be categorized into two based on whether a lossless reference image is available: full-reference (FR-IQA)\cite{6562812, 7934456, 6247789, 6894484} and no-reference (NR-IQA). Unlike FR-IQA, NR-IQA\cite{6272356, 6172573, 7094273} does not have access to reference images which is harder and more widely used. With the success of deep learning and convolution neural networks (CNNs) in computer vision tasks, CNN-based methods have significantly outperformed traditional approaches in handling real-world distortions. Early deep learning NR-IQA methods employed stacked CNNs for feature extraction. Ma\cite{8110690} proposes a multi-task network where two sub-networks are trained for distortion identification and quality assessment. Hyper-IQA\cite{9156687} leverages both low-level and high-level features and makes the latter redirect the former. Furthermore, Zhu\cite{9156932} proposes a model that employs meta-learning to capture shared prior knowledge among different distortions. But in the era of high-resolution Image Quality Assessment, both methods are faced with substantial computational challenges while processing entire images. There already exist some data-efficient methods for quality assessment, such as FAST-VQA\cite{Wu2022FASTVQAEE} with random-crop. But cropping cannot guarantee that the remained part matters. 

Compressive sensing is a novel method that breaks through the limitations of the Nyquist theorem on signal sampling\cite{1614066, 4472240}. This technique shows that one can faithfully reconstruct an entire image from a minimal amount of measurements compared to the original image, thus significantly simplifying the sampling process.  

The concept of compressive learning (CL) was first proposed by Calderbank et al.\cite{article} and Davenport et al.\cite{10.1117/12.714460} when they built the inference system directly using measurements without reconstruction. Tran\cite{9070152} makes some theoretical works and gradually finds some multidimensional properties of CL. A recent work \cite{9841016} by Chong Mou et al. employs transformer as the backbone to cope with information loss through element-wise correlations and achieves amazing performance. 

In many applications, such as NR-IQA, the precise reconstructed image isn't a primary concern. Still, we emphasize more on the final score, which means we can directly perform high-level tasks after initialization or even on the compressed domain.Inspired by this work, to make NR-IQA more data-efficient and extract the feature of the original image globally, we introduce compressive sensing in NR-IQA task.We propose CL-IQA that integrates compressive learning with the NR-IQA task and CS-IQA which skip the AEM block of CL-IQA. They share the same backbone. Firstly, CSM allow us obtain measurements at an arbitrary ratio. Secondly, the measurements are adaptive embedded by AEM. Thirdly, the measurements or embedded measurements were fed into the vision transformer and SSTM to extract deep features. Finally, a dual-branch structure is employed for the score prediction. Our main contributions are as follows: 

\begin{figure}
    \centering
    \includegraphics[width=1\linewidth]{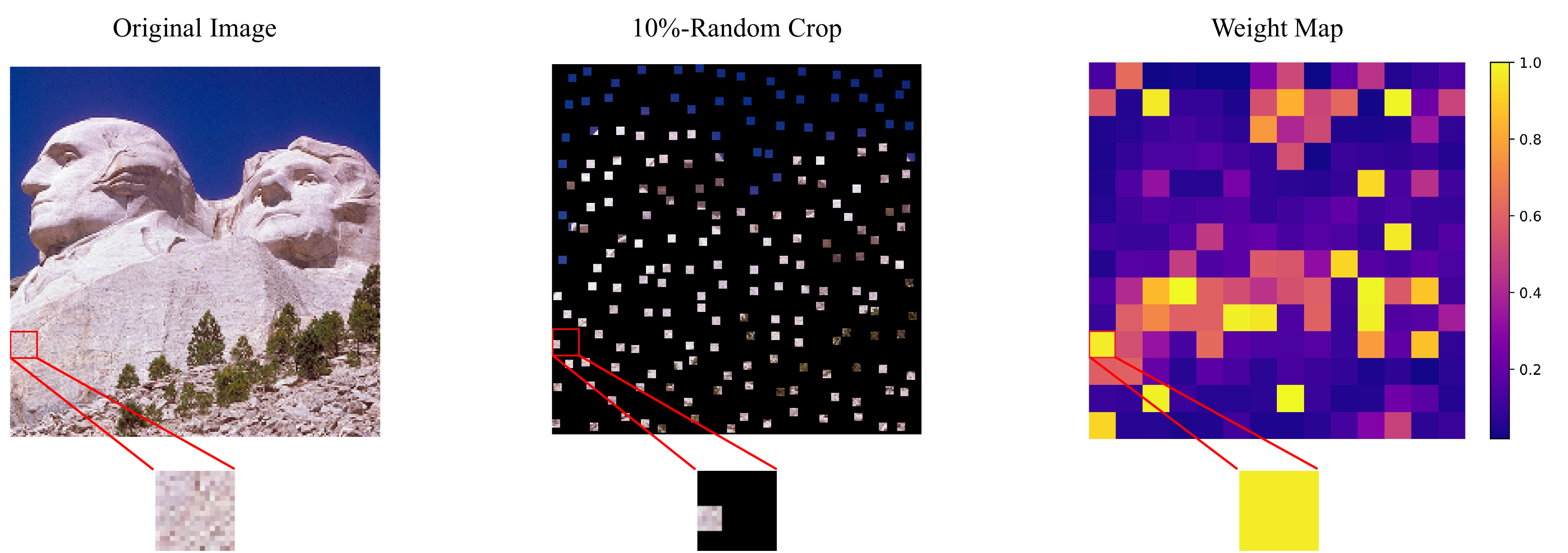}
    \caption{Random-crop removes pixels from every block despite the importance, leading to severe information loss.}
    \label{fig:figure2}
\end{figure}

$\bullet$ We apply compressive learning to image quality assessment with, enable sampling of images at arbitrary ratios. To our knowledge, CL-IQA is the first work integrating compressive learning and image quality assessment.

$\bullet$ We pretrain the CSM with lightweight compressive sensing network enable it captures the feature about quality assessment more effectively. The proposed CS-IQA significantly reduce the complexity by skip the AEM module.

$\bullet$ Experiments show that CL-IQA and CS-IQA outperforms state-of-the-art IQA methods with benefiting from the sampling modules.

\begin{figure*}[t]
	\centering
	\includegraphics[width=1\linewidth]{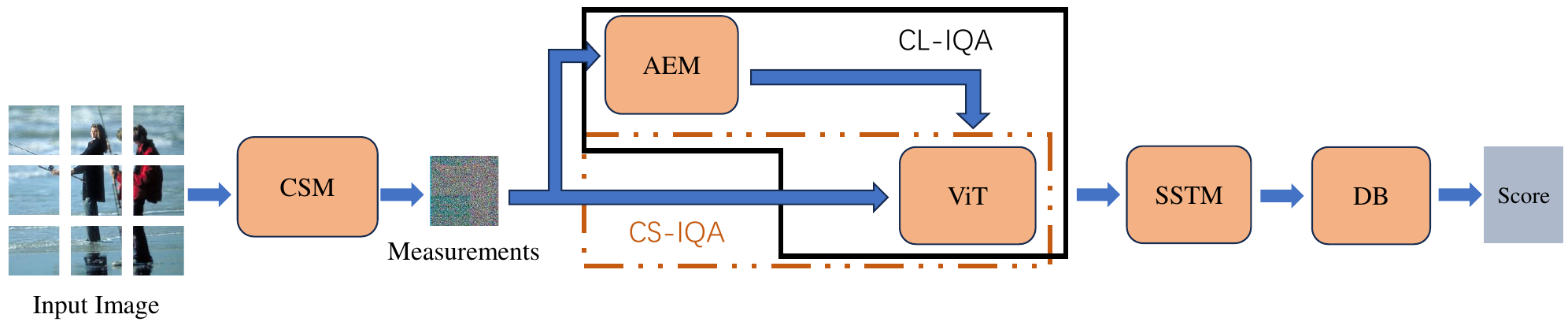}
	\caption{The architecture of our proposed CS-IQA and CL-IQA.}
	\label{fig:figure1}
\end{figure*}

\section{Proposed Approach}

In this section, we elaborate on the details of our framework and integrated model. The whole architecture is presented in Fig.\ref{fig:figure1}, which is mainly composed of a Compressed Sampling Module (CSM), a Adaptive Embedding Module (AEM), a Vision Transformer, a Scale
Swin TranBlocksformer Moudle (SSTM) and a Dual Branch (DB).

\subsection{Compressed Sampling Module}

In Fig.\ref{fig:figure2}, we show the original image, 10\%-cropped image, and the weight map of the original one. We can observe that even for the block with a large weight, random-crop removes 90\% pixels of it. Actually, instead of the remove pixels randomly, we might only need a small number of measurements sampled globally from original image (perhaps 25\% measurements or even less) to achieve highly reliable score prediction. In implementation, we employ compressive sensing for sampling.

CSM is employed to compress the image using compressive sensing with sampling matrix $\Phi$, which is learnable. Due to the high computational complexity, previous sampling applications are limited in datasets with low-resolution images. To extend the model's applicability to real-world images, we use block-based compressive sensing algorithm (BCS)\cite{9159912}. BCS is able to compress the image block-by-block. As shown in Fig.\ref{fig:figure1}, given an image of size $H\times W$, we split the image into $L=\frac{H}{B}\times \frac{W}{B}$ non-overlapping blocks, with each block sized $B\times B$. To get the measurements, we have a base sampling matrix $\Phi$ sized $B^2 \times B^2 $. Noting $\gamma$ as the sampling ratio, we truncate the first $\gamma B^2$ rows of $\Phi$ to get $\Phi_{\gamma}$. Finally, to obtain the measurements, the process can be described as follows:
\begin{equation}
	y_i = \Phi_{\gamma}\cdot x_i
\end{equation}
where $x_i \in \mathbf{R}^{B^2}$ represents the flattened block of original image, $y_i \in \mathbf{R}^{\gamma B^2}$ denotes the corresponding measurement, and $i \in [1,L]$. Throughout the whole process, the only parameter is $\Phi$. To accelerate the computation, we can treat every row of $\Phi_{\gamma}$ as a convolution kernel, which turns matrix multiplication into convolution operation.

There are many CS methods but some of them have extremely high computational complexity due to the iterative calculations, such as ISTA-Net\cite{zhang2018ista},AMS-Net\cite{9855869} ,OPINE-Net\cite{zhang2020optimization}. Shi et al.\cite{8019428} propose a lightweight network model called CSNet and show its promising performance. Therefore we pretrain our CSM with CSNet.

\begin{figure}
    \centering
    \includegraphics[width=1\linewidth]{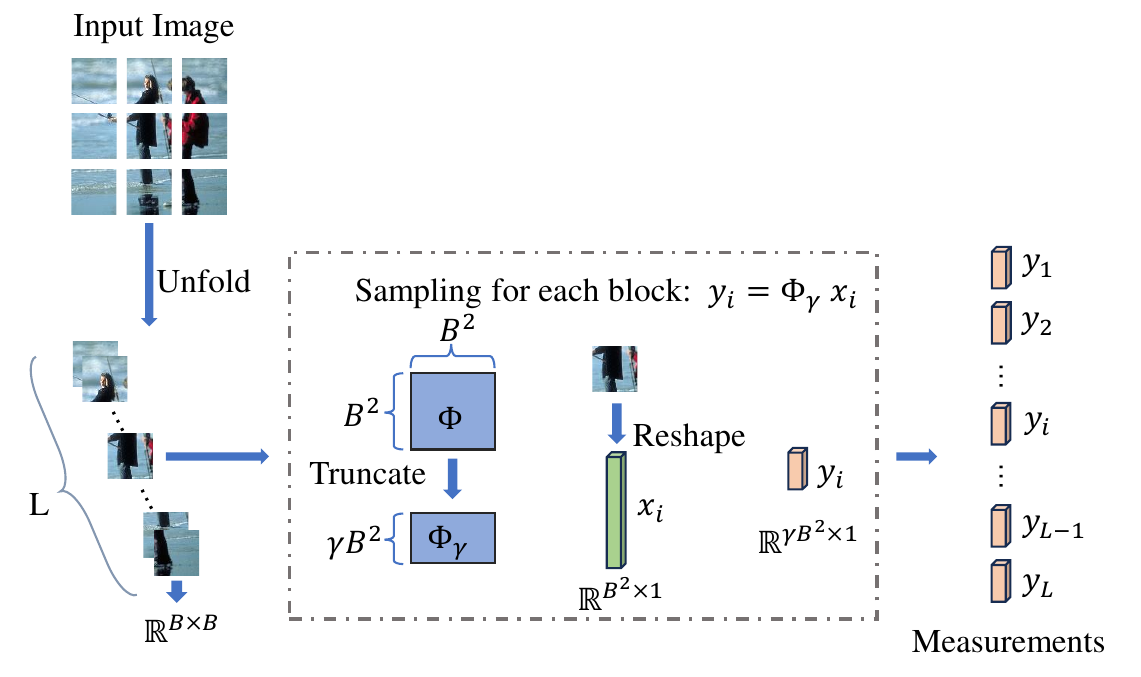}
    \caption{Process of Flexible Sampling Module (AEM).}
    \label{fig:figure3}
\end{figure}

\subsection{Adaptive Embedding Module}

Considering that IQA demands a global perception of the whole image, we employ transformer for its excellent performance in constructing long-range correlations.  Similar to pure transformer, ViT embeds the input image into the size of $L\times d$, where $d$ denotes the embedding dimension. However, due to the arbitrary CS ratio, the input sequence size $L\times \gamma B^2$ of CL-IQA is also variable according to the parameter $\gamma$. As shown in Fig.\ref{fig:figure3}, to overcome the limitation of the fixed embedding module in ViT, we have a base embedding matrix M sized $B^2 \times B^2$. Then, according to the given ratio $\gamma$, we use the first $\gamma B^2$ columns of $M$ to form $M_{\gamma}$. Finally, for every measurement $y_i$, the embedding process is as follow:
\begin{equation}
	t_i = M_{\gamma}\cdot y_i
\end{equation}
Where $t_i \in \mathbf{R}^{d}$ represents the sequence that can be handled by ViT. 

\subsection{Vision Transformer}

Noting $T=[t_1,t_2,…,t_L]$ and $P$ as the positional embedding result, the initial input of ViT is constructed by the following formulation:
\begin{equation}
	X^0 = T + P
\end{equation}

The standard ViT\cite{DBLP:journals/corr/abs-2010-11929} has the same structure as the transformer encoder, containing a stack of $N$ consecutive blocks. Within each block, the attention mechanism is employed to compute correlations between image patches, finally giving high-level features. The computation process is as follows:
\begin{eqnarray}
	\tilde{X}^i &=& Norm(MSA(X^{i-1}) + X^{i-1})\\
	X^i &=& Norm(FF(\tilde{X}^i) + \tilde{X}^i)
\end{eqnarray}
Where $MSA$ represents multi-head self-attention, $Norm$ denotes normalization, $FF$ indicates feed forward, $X^i$ is the output of the $i$-th block and $i \in [1,N]$.

\subsection{Scale Swin TranBlocksformer Moudle}

The Scale Swin Transformer Moudle consists of Swin Transformer Layers\cite{Liu_2021_ICCV} and a convolutional layer. The SSTM first encodes the input feature through 2 layers of STL: 
\begin{equation}
	X_{1}^i = S_{STL}(X_{0}^i)
\end{equation}
Then the convolutional layers are applied before the residual connection. The output of SSTB is formulated as:
\begin{equation}
	X_{1}^i = \alpha \cdot S_{CONV}(X_{1}^i) + X_{1}^i
\end{equation}
the coefficient $\alpha$ denotes the scale factor of the output of 2 layers of STL.

\begin{figure}
    \centering
    \includegraphics[width=1\linewidth]{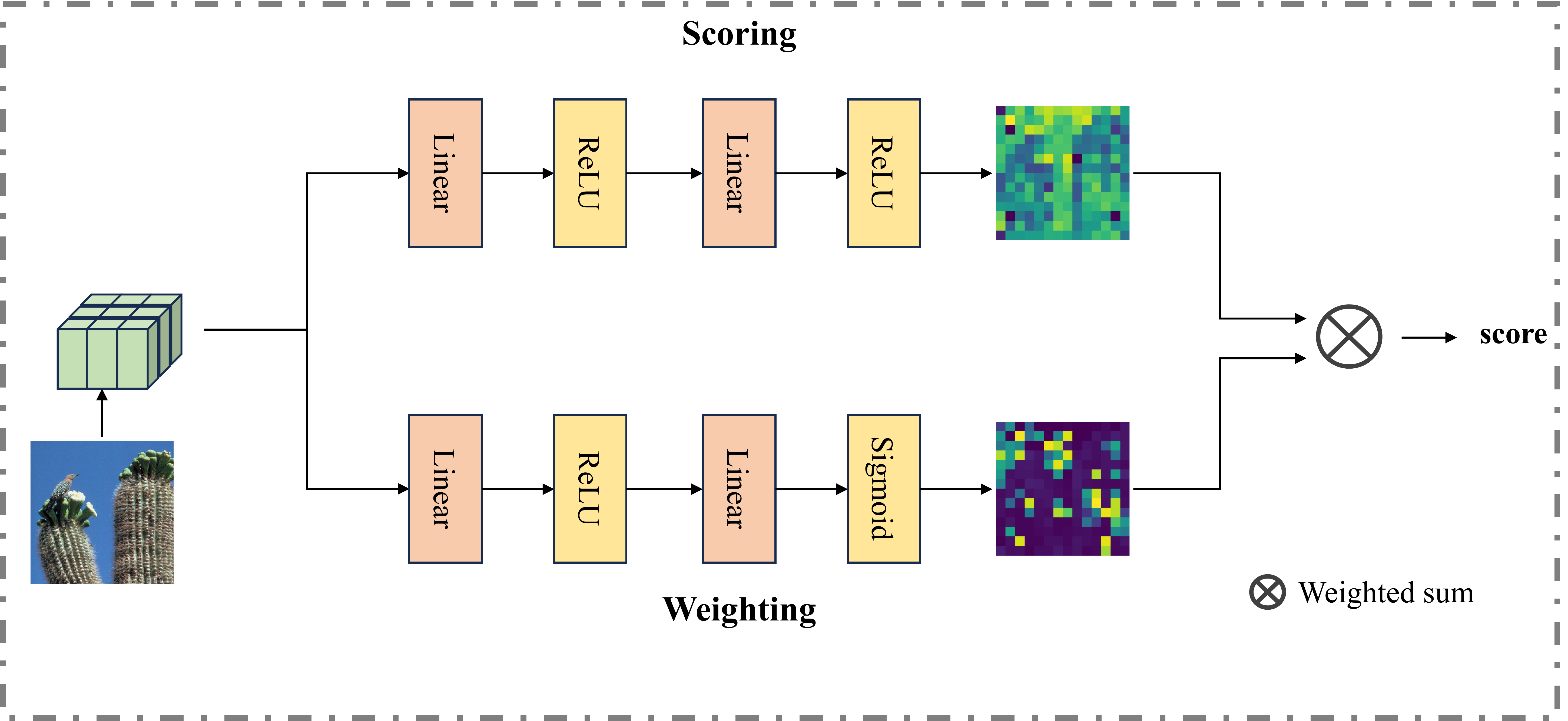}
    \caption{The dual-branch (DB) structure. Each branch contains two fully connected layers for the score and weight prediction.}
    \label{fig:figure4}
\end{figure}

\subsection{Dual Branch}

For the final scoring, most of recent IQA methods \cite{9156687}, \cite{9156932}, \cite{9706735}, \cite{Saha2023ReIQAUL} utilize MLP, i.e., single branch. Nevertheless, when we give a final score of a patch in the image, the clarity is not the only thing of importance, some other factors like aesthetics or the content also matter, which we call weight. To combine the two elements, we design the dual-branch structure for final quality score prediction. As shown in Fig.\ref{fig:figure4}, the whole module consists of two independent branches: a scoring branch and a weighting branch. Both branches are rather simple with the same structure except for the last activation function. Given the feature $F=X^N$, the final score is obtained by the formula:
\begin{equation}
	score = \frac{Sum(S(F)\times W(F))}{Sum(W(F))}
\end{equation}
where $S$ indicates scoring branch and $W$ is weighting branch.

\begin{table*}[t]
    \centering
    \caption{Comparisons with state-of-the-art NR-IQA algorithms.}
    \resizebox{0.85\linewidth}{!}{
    \begin{tabular}{ccccccccc}
    
    \hline
         &  \multicolumn{2}{c}{LIVE}&  \multicolumn{2}{c}{CSIQ}&  \multicolumn{2}{c}{TID2013}&  \multicolumn{2}{c}{KADID-10K}\\
         Method&PLCC&SRCC&PLCC&SRCC&PLCC&SRCC&PLCC&SRCC	\\
         \hline
         
         WaDIQaM\cite{8063957}            &  0.955                        &  0.960                        &  0.844                        &  0.852                        &  0.855                        &  0.835                        &  0.856                        &0.851                        \\
         DBCNN\cite{8576582}              &  0.971                        &  0.968                        &  0.959                        &  0.946                        &  0.865                        &  0.816                        &  0.855                        &0.850                        
\\
         TIQA\cite{9506075}               &  0.965                        &  0.949                        &  0.838                        &  0.825                        &  0.858                        &  0.846                        &  0.755                        &0.762                        
\\
         MetaIQA\cite{9156932}            &  0.959                        &  0.960                        &  0.908                        &  0.899                        &  0.868                        &  0.856                        &  0.849                        &0.840                        
\\
         P2P-BM\cite{9157380}             &  0.958                        &  0.959                        &  0.902                        &  0.899                        &  0.856                        &  0.862                        &  0.845                        &0.852                        
\\
         HyperIQA\cite{9156687}           &  0.966                        &  0.962                        &  0.942                        &  0.923                        &  0.858                        &  0.840                        &  0.858                        &0.915                        
\\
         TReS\cite{9706735}               &  0.968                        &  0.969                        &  0.942                        &  0.922                        &  0.883                        &  0.863                        &  0.885                        &0.872                        
\\
         Re-IQA\cite{Saha2023ReIQAUL}             &  0.971                        &  0.970                        &  0.960                        &  0.947                        &  0.861                        &  0.804                        &  0.946  &0.944 
\\
         MANIQA\cite{Yang2022MANIQAMA}             &  0.983 &  0.982 &  0.968                        &  0.961                        &  0.943                        &  0.937 &  0.885                        &0.872                        \\
    \hline
 CL-IQA-10& 0.944                        & 0.936                        & 0.971                        & 0.965                        & 0.930                        & 0.910                        & 0.938                        &0.933                        
\\
 CL-IQA-20& 0.975                        & 0.970                        & 0.978  & 0.970  & 0.947  & 0.925                        & 0.942                        &0.937                        
\\
 CL-IQA-50& \textbf{0.984} & \textbf{0.979} & \textbf{0.983} & \textbf{0.975} & \textbf{0.954} & \textbf{0.944} & \textbf{0.948} &\textbf{0.943} 
\\
 CS-IQA& 0.947& 0.944& 0.973& 0.965& 0.940& 0.926& 0.945&\textbf{0.943} \\
 \hline
    \end{tabular}}
    \label{tab:SOTA_table}
\end{table*}

\begin{table}
    \centering
\caption{ ratio-fixed models perform best under the training CS ratio, while CL-IQA-r is rather stable.}
\label{tab:my_label1}
    \begin{tabular}{ccccc}
    \hline
         \multirow{2}{*}{CS Ratio}&  \multicolumn{4}{c}{PLCC/SRCC}\\
         &  CL-IQA-10             &  CL-IQA-20             &  CL-IQA-50             & CL-IQA-r              \\
         \hline
         10\%                      &  \textbf{0.971/0.965} &  0.571/0.550          &  0.276/0.288          & 0.962/0.936          
\\
         20\%                      &  0.688/0.629          &  \textbf{0.978/0.970} &  0.772/0.765          & 0.969/0.953          
\\
         50\%                      &  0.655/0.606          &  0.822/0.753          &  \textbf{0.983/0.975} & 0.977/0.972          
\\
         100\%                     &  0.751/0.706          &  0.834/0.780          &  0.893/0.838          & \textbf{0.976/0.970} \\
         \hline
    \end{tabular}
\end{table}

\begin{table}
    \centering
\caption{Ablation results of ViT and DB at CS ratio=10\%.}
\label{tab:my_label2}
    \begin{tabular}{cccccc}
    \hline
 \multicolumn{2}{c}{Module}& \multicolumn{4}{c}{PLCC / SRCC
}\\
         ViT           &DB          &  LIVE                 &CSIQ                 &TID2013              &KADID-10K            \\
         \hline
         \XSolidBrush&  \XSolidBrush&  0.901/0.890          &  0.946/0.930          &  0.897/0.868          & 0.917/0.911          
\\
         \Checkmark&  \XSolidBrush&  0.938/0.934          &  0.966/0.952          &  0.921/0.897          & 0.933/0.926          
\\
         \Checkmark&  \Checkmark&  \textbf{0.944/0.936} &  \textbf{0.971/0.965} &  \textbf{0.930/0.910} & \textbf{0.938/0.933} \\
    \hline
    \end{tabular}

\end{table}

\section{Experiments}

\subsection{Experimental Settings}

We implement our experiments on NVIDIA GeForce GTX 1080Ti with PyTorch 1.10.0 and CUDA 11.3 version for the whole training, validating, and testing process. We train and evaluate the performance of our proposed model on four datasets: LIVE\cite{1709988}, CSIQ\cite{Larson2010MostAD}, TID2013\cite{Ponomarenko2016ImageDT}, KADID-10K\cite{8743252}. And T91 dataset is used to pretrain the CSM.

We choose ViT-B/8-224\cite{DBLP:journals/corr/abs-2010-11929} as our backbone which is pre-trained on ImageNet-21k with patch size set to 8 and image size 224. This base version transformer contains $N$=12 transformer blocks, with the number of heads being 12 in each layer and the dimension of the feature embedding is set to 768. To be consistent with the patch size used in transformer and protect local information correlations, we set the block size in the sampling model also to 16.

In the experiments, we randomly split the dataset into 8:2 for training and testing. During training, we set the batch size to 8. We utilize Adam as the optimizer with learning rate $1\times 10^{-5}$ and weight decay $1\times 10^{-5}$. The training loss we use is Mean Square Error (MSE) loss. During testing, we select the one that performs the best in the validation dataset. In CL-IQA, for each image in the test dataset, we randomly crop a $224 \times 224$ sized patch of it for 5 times and obtain the final score by averaging the results. In our experiments, we evaluate our CL-IQA with both fixed CS ratios and arbitrary ratios. For instance, “CL-IQA-10” represents the variant with a fixed CS ratio of 10\%, and “CL-IQA-r” denotes the variant that deals with an arbitrary CS ratio.

\subsection{Comparison with State-of-the-Art Methods}

Table \ref{tab:SOTA_table} shows the overall performance of CL-IQA with fixed sampling ratio (10\%, 20\%, and 50\%) and CS-IQA on four standard datasets in terms of PLCC and SRCC. We can observe that compared with the methods that utilize the whole image to predict, our proposed methods achieves state-of-the-art performance with much less data. Even if only 10\% measurements are available in CSIQ, we can obtain excellent outcomes. Furthermore, one can see that for the same dataset, the performance is getting better with the CS ratio growing.

\begin{figure}
    \centering
    \includegraphics[width=1\linewidth]{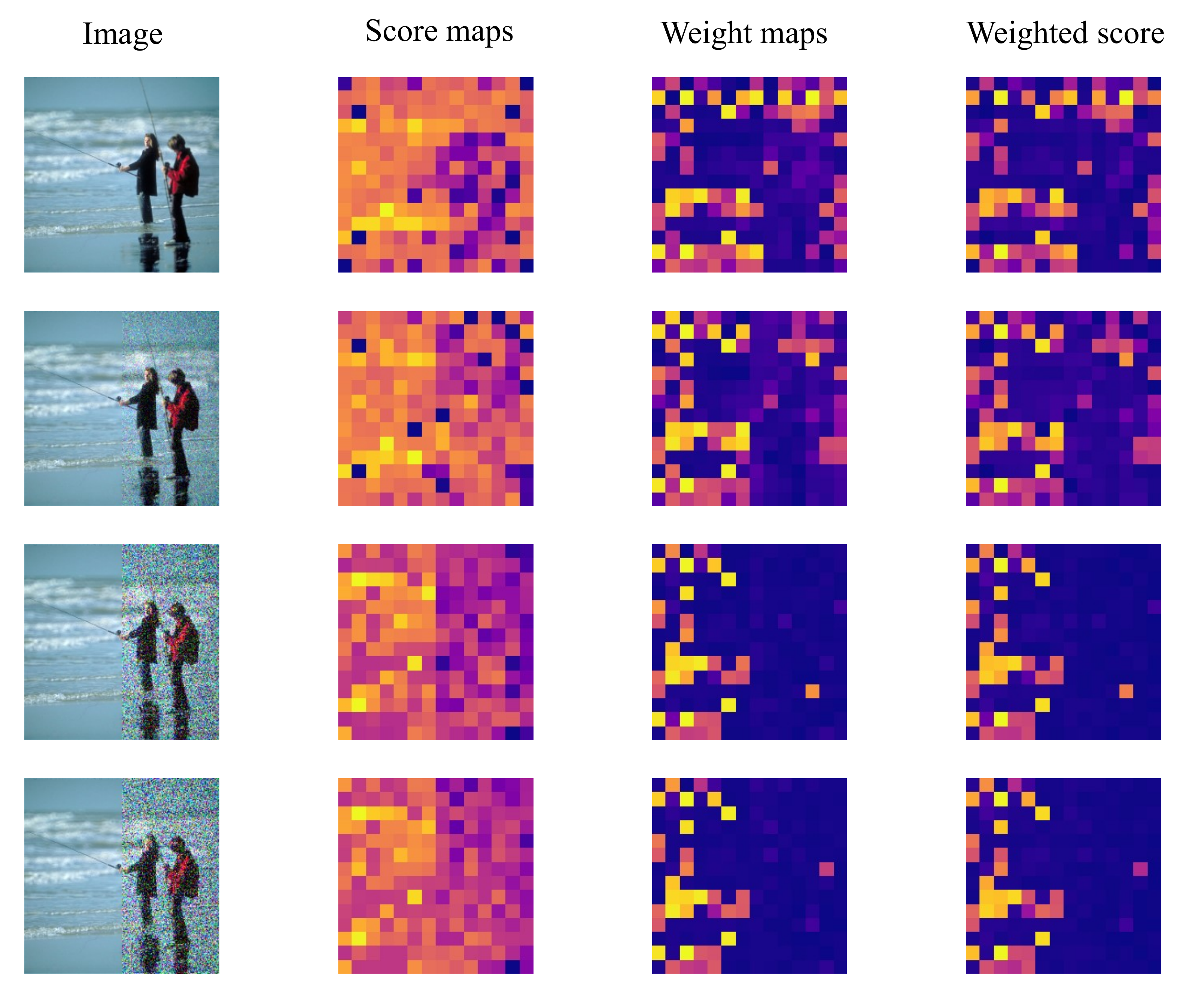}
    \caption{The dual-branch (DB) structure. Each branch contains two fully connected layers for the score and weight prediction.}
    \label{fig:figure5}
\end{figure}

\subsection{Visualization}

To figure out exactly how the dual-branch structure helps when predicting the quality score, we generate different feature maps using CL-IQA-10. As shown in Fig.\ref{fig:figure5}, we add White Gaussian Noise to the right half of the image according to different SNRs(Signal-to-Noise Radio) (10, 1, and 0.1). As what we have assumed, different patches of the image matter differently and the map tends to give more weight to the outline or background. 

\subsection{Ablation Study}

In this section, we conduct ablation study on CL-IQA-10 to prove the effectiveness of the transformer backbone and the dual branch. For the transformer part, we use ResNet-101 to replace it for feature extraction. In the research of dual branch (DB), we replace it with two layers of CNN and MLP. The result is shown in Table \ref{tab:my_label1}. We can observe that each part has a certain effect on the quality prediction. Transformer has the superiority of constructing long-range correlations and the dual-branch can reasonably allocate weight among the patches.

\section*{CONCLUSION}

In this paper, we propose a new data-efficient framework for NR-IQA dubbed CL-IQA and CS-IQA. It samples the image using the Compressed Sampling Module to obtain measurements globally. Then we utilize the adaptive embedding to overcome the problem that measurements don’t match the fixed input shape of the transformer. The measurements are fed into the Vision Transformer and Scale Swin TranBlocksformer Moudle for further feature extraction. Finally, the DB weights every patch of the image and predicts the final quality score. Experiments demonstrate that our proposed methods outperforms many recent state-of-the-art NR-IQA methods.

% \bibliographystyle{IEEEbib}
% \begin{thebibliography}{00}
% \bibitem{b1} Ma K, Liu W, Zhang K, Duanmu Z, Wang Z, Zuo W. End-to-End Blind Image Quality Assessment Using Deep Neural Networks. IEEE Trans Image Process. 2018 Mar;27(3):1202-1213

% \end{thebibliography}

\bibliographystyle{IEEEbib}
\bibliography{S-IQA}

\end{document}